\pdfoutput=1

\documentclass[11pt]{article}

\usepackage[]{naacl2021}

\usepackage{times}
\usepackage{latexsym}

\usepackage[T1]{fontenc}

\usepackage[utf8]{inputenc}

\usepackage{microtype}

\usepackage{pgfplots}
\usepackage{tikz}
\usepackage{collcell}
\usepackage{booktabs}
\usepackage{wrapfig}
\usepackage{multicol}
\usepackage{comment}
\usepackage{multirow}
\usepackage{listings}
\usepackage{enumitem}
\usepackage{color, colortbl}
\definecolor{Gray}{gray}{0.9}

\usepackage[first=0,last=9]{lcg}

\usepackage{pgfplots}
\usepackage{tikz}
\usepackage[compat=1.0.0]{tikz-feynman}
\usepackage{tikz}
\usepackage{collcell}
\usepackage{caption} 
\usepackage[acronym]{glossaries}
\usepackage{comment}
\usepackage{graphicx}
\usepackage{xcolor}
\usepackage[ruled,vlined]{algorithm2e}
\usepackage{float}
\usepackage{mathtools}
\usepackage[acronym]{glossaries}
\usepackage{booktabs} 
\usepackage{tikz}
\usepackage{amsmath}
\usetikzlibrary{fit,positioning,arrows}
\usetikzlibrary{shapes,arrows}
\usetikzlibrary{intersections}
\usepackage[T1]{fontenc}
\usepackage{caption}
\usepackage{booktabs}  
\usepackage{siunitx}

\newacronym{set}{SET}{Search at Every Token}
\newacronym{slt}{SLT}{Search at Last Token}
\newacronym{ss}{SS}{Skip Stop-words}
\newacronym{sm}{SM}{Similarity Matching pre-trained model}
\newacronym{dqn}{DQN}{Deep Q Networks}
\newacronym{ts}{TS}{Average Number of Triggered Searches}
\newacronym{delay}{Effort}{Average Effort}
\newacronym{theta}{\Theta}{Increase in delay for every saved query}

\tikzset{
  main/.style={circle, minimum size = 5mm, thick, draw=black!80, node distance = 10mm},
  connect/.style={-latex, thick},
  box/.style={rectangle, draw=black!100}
}

\title{Deep Reinforcement Agent for\\ Efficient Instant Search}

\author{Ravneet Singh Arora \\
  Bloomberg, USA \\
  \texttt{rarora62@bloomberg.net} \\\And
  Sreejith Menon\thanks{This work was done as an employee of Bloomberg} \\
  Google, USA \\
  \texttt{sreejithmenon@google.com} \\\AND
    Ayush Jain  \\
  Bloomberg, USA \\
  \texttt{ajain448@bloomberg.net} \\\And
     Nehil Jain \\
  Bloomberg, USA \\
  \texttt{njain42@bloomberg.net}}

\begin{document}
\maketitle
\begin{abstract}
Instant Search is a paradigm where a search system retrieves answers on the fly while typing. The naïve implementation of an Instant Search system would hit the search back-end for results each time a user types a key, imposing a very high load on the underlying search system. In this paper, we propose to address the load issue by identifying tokens that are semantically more salient towards retrieving relevant documents and utilize this knowledge to trigger an instant search selectively. We train a reinforcement agent that interacts directly with the search engine and learns to predict the word's importance. Our proposed method treats the underlying search system as a black box and is more universally applicable to a diverse set of architectures. Furthermore, a novel evaluation framework is presented to study the trade-off between the number of triggered searches and the system's performance. We utilize the framework to evaluate and compare the proposed reinforcement method with other intuitive baselines. Experimental results demonstrate the efficacy of the proposed method towards achieving a superior trade-off. 

\end{abstract}

\maketitle

\section{Introduction}
\begin{figure*}
\includegraphics[width=\textwidth]{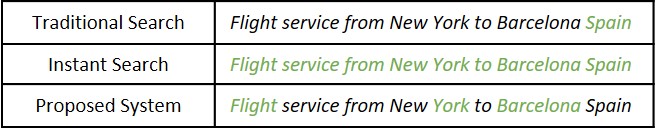}
\caption{Behaviors of Traditional, Instant and proposed Instant search system for the query \textit{Flight service from New York to Barcelona Spain}. Searches are triggered at tokens marked green. The whole prefix is forwarded to the search engine as a query.}

\label{fig:sayt-main}
\end{figure*}

Interactivity in search engines has substantially grown in popularity in recent years. To further enrich the user experience, most modern search engines such as Google and Bing provide instant search capabilities \cite{venkataraman2016instant}. Instant search retrieves results on the fly at every keystroke compared to conventional search engines that trigger search at the end of the query. Analyses of query-logs performed by \cite{cetindil2012analysis} have shown that the instant search improves user experience by reducing the overall time and effort to retrieve the relevant results and helps users find information when they are not sure of the exact query. This feature is very relevant to mobile applications. Recently, these systems have also become extremely popular in Social Networking websites such as Linkedin \cite{venkataraman2016fast}.  Instant Answers is another variation of this paradigm, which is very common in search engines these days. Instant answers allows users to view answers instantly while typing questions such as \textit{``how is weather today?''}
This feature is also handy in open-domain question answering where user needs are ambiguous.

The implementation of instant search systems faces a significant challenge in the form of immense load on the back-end search engine. The instant search leads to an increase of tens or up to hundreds of more queries for a single search session. This becomes more severe in the case of longer natural language queries. Managing such load becomes problematic for several reasons: the software or the hardware might not be able to cope with high query throughput during spikes of requests, or it might cause high energy consumption by the servers, or just consume computational resources needed by other processes like indexing.

Several approaches have been proposed to improve the performance and scalability of instant search. Many of these studies are based on designing more efficient index data structures for faster retrieval of results~\cite{Bast2006TypeLF,Fafalios2011ExploitingAM,Li2012SupportingET,ji2009efficient,Li2011EfficientFF,Wang2010InteractiveAF}. These data structures are examined together with the techniques such as caching ~\cite{Fafalios2012ScalableFA} for their ability to improve the search engine query throughput. Caching has been further extensively applied for large-scale traditional search systems in various studies such as ~\cite{MARKATOS2001137,saraiva2001rank,dean2009challenges,gan2009improved,fagni2006boosting,long2006three}. New index data-structures and file system formats for boosting the overall speed of search engines have also been explored ~\cite{brin1998anatomy,dean2009challenges}.  

In this paper, we propose a new method to solve the instant search paradigm's scalability challenges. Our approach stems from the idea that a subset of tokens heavily influences the retrieval of the most relevant results. This subset generally includes keywords that are either topical or tokens that can alter the semantic meaning of the query. 
We have applied this idea towards training a reinforcement agent that predicts if a typed token is salient and selectively triggers search only for such tokens. This is illustrated in Figure \ref{fig:sayt-main}. 
Searches are triggered at tokens marked green. A traditional search system would wait till the last token before issuing the search; an instant search system, on the other hand, queries at every new token. Our proposed approach, in addition to common stopwords, decides to skip the search at word \texttt{New} as it is very common and needs more context (\texttt{York} in this case) to retrieve the correct answer. Also, since there is only one Barcelona city present in Spain, the word \texttt{Spain} does not influence the results returned and hence is skipped. The new approach treats the underlying retrieval engine as a black box and is decoupled from the internal implementation. During the training, the agent updates weights based on the feedback received during its interaction with the search system. This methodology has the following advantages: a) More universal application to a diverse set of modern architectures; b) No need to scale up individual components of complicated search and QA pipelines such as \cite{yang2019end}; c) Easy integration with the existing techniques such as caching.
Reinforcement learning provides the framework to integrate and experiment with different reward functions. Furthermore, there can be a lot of different states based on the decision taken by the algorithm and it is not easy to calculate exact true labels for a pure supervised setting. Recently, reinforcement learning has been successfully applied to an identical problem in the field of Simultaneous Machine Translation (SMT)~\cite{grissom,satija2016simultaneous,Neubig2017LearningTT}. SMT is defined as the task of producing a real-time translation of a sentence while typing. The goal here is to achieve a good trade-off between the quality and delay of the translation. 

We further evaluate the loss in the quality of instant search due to introducing the proposed reinforcement agent. Instant search quality is measured based on the studies that have compared the instant search system with a traditional one ~\cite{cetindil2012analysis,chandar2019developing}. Instant search query logs have been analyzed by ~\cite{cetindil2012analysis} to understand the properties of instant search that lead to a better user experience. Recently \cite{chandar2019developing} combined user-query interaction logs with user interviews and proposed new metrics that can evaluate user satisfaction for an instant search system. Both the studies have proposed results-quality and user-effort as the two primary metrics to measure user experience improvement. Quality measures how relevant the search system results are to the user query, whereas Effort captures how quickly the relevant results are retrievable using a search engine. We use these metrics to estimate how well the proposed methods can reduce the overall system load while preserving the performance. Experiments are performed on three different combinations of datasets with two retrieval systems. Our experiments show that the proposed model achieves a superior trade-off by achieving near-optimal performance while reducing the number of triggered searches by 50\%.


\section{Baselines}
This section introduces the baselines that are evaluated and compared with the proposed model.

\noindent \textbf{Search at Last Token} \label{SET}: SET issues search for every new token. This method represents the true instant search paradigm.

\noindent \textbf{Search at Last Token}\label{SLT}: SLT waits for the entire query and triggers a single search request at the end. This baseline mimics the behavior of a regular retrieval engine. 

\noindent \textbf{Skip Stop-words}: Skip Stop-words simply issues a search at every token except the stop-words.

\noindent \textbf{Similarity Matching pre-trained model}: 
Similarity Matching pre-trained model issues a query only when the query's semantic meaning has changed by more than a certain threshold.
We utilize the pre-trained  Universal Sentence Encoder model \cite{cer2018universal} to generate an embedding for the query at every new token and compare the similarity with the embedding of the previously searched sub-query. We use $CosineDistance$ between sentence embedding vector pairs to measure the similarity. A sentence pair $S_1, S_2$ is considered to be semantically different if $CosineDistance$($S_1, S_2$) $\geq$ $threshold$ \cite{gomaa2013survey}. We treat the $threshold$ as a hyper-parameter, and the actual value is later stated in Section \ref{ssec:eval}. Algorithm \ref{algosemsim} describes this approach in more detail.



\begin{algorithm}[]
\SetAlgoLined
\SetKwData{Threshold}{threshold}
\SetKwData{Previous}{previous}
\SetKwFunction{Current}{current}
\SetKwFunction{CosineDistance}{CosineDistance}
\SetKwFunction{RetrieveDocuments}{RetrieveDocuments}
\SetKwFunction{GetEmbedding}{GetSentenceEmbedding}
 $Q \leftarrow$ Query \;
 $N \leftarrow$ Number of tokens in query Q \;
 $D \leftarrow$ Set of Retrieved Documents \;
 $q_{searched} \leftarrow$ Sequence of tokens previously searched \;
 $V_{searched} \leftarrow$ Embedding Vector of $q_{searched}$ \;
 $q_{current} \leftarrow$ Current sequence of tokens \;
 $V_{current} \leftarrow$ Embedding Vector of $q_{current}$ \;
 \For{$i\leftarrow 1$ \KwTo $N$}{
    $q_{current} \leftarrow$ $Q[1,i]$\;
    $V_{current} \leftarrow$ \texttt{GetEmbedding($q_{current}$)}\;
    \BlankLine
    \If{\texttt{\CosineDistance($V_{searched}$, $V_{current}$)} $\geq$ $threshold$}
    {
        $q_{searched} \leftarrow q_{current}$ \;
        $V_{searched} \leftarrow V_{current}$ \;
        $D \leftarrow$ \RetrieveDocuments{$q_{current}$}\;
    }
 }
\caption{Inference using Similarity Matching pre-trained model Method}\label{algosemsim}
\end{algorithm}

\section{Reinforcement Agent}

\noindent \textbf{Deep Q Networks}:
In Q-learning~\cite{watkins1992q}, the environment is formulated as a sequence of state transitions $(s_t, a_t, r_t, s_{t+1})$ of a Markov Decision Process (MDP). 
At a given time-step $t$ for state $s_t$, the agent takes an action $a_t$ and in response receives the reward $r_t$. 
As a result, the environment transitions into state $s_{t+1}$.
The agent chooses action $a_t$ for the state $s_t$ by referring a state-action value function $Q(s_t, a_t)$, which measures the action’s expected long-term reward.
The algorithm updates the Q-function by interacting with the environment and obtaining rewards. In large environments, it is impractical to maintain a $Q$ function for a substantially large number of states. DQN~\cite{Mnih2013PlayingAW} solves this problem by approximating  $Q(s,a)$ using a deep neural network, which takes state $s$ as input and calculates value for every state/action pair.


\noindent \textbf{Environment}:
The environment yields new words for the agent and also interacts with the underlying search engine. For a given query, the agent receives a new word $x_t$  from the environment at every time-step $t$ and, in response, takes action $a_t$. 
Based on the action, the environment requests the underlying retrieval engine, and the agent is provided feedback in the form of reward $r_t$. 
An episode terminates at the last token $x_T$ of the query.

\noindent \textbf{State}\label{sec:state}:
The state represents the portion of the query that is already observed by the Agent. 
For a given query $q$, let us assume that the agent has received tokens $x_1,\cdots,x_t$ denoted by partial query $q'$. 
The environment maintains two sequences of tokens for every $q'$:
\begin{itemize}
\item $q'_1$: the list of tokens  $x_1,\cdots,x_{t'}$ used in the last search query submitted to the system.
\item $q'_2$: the list of tokens $x_{t'+1}\cdots,x_t$ the system has seen since it last submitted a search query.
\end{itemize}

This state formulation allows the agent to learn the overall importance of $q'$ conditioned on already searched sequence $q'_1$. 
At every time-step $t$, the agent receives a new token $x_t$ which is then appended to the unseached sub-sequence $q'_2$: $q'_2$ = $q'_2 \cup x_t$. After a search is triggered, $q'_2$ is appended to the searched sub-sequence $q'_1$ and $q'_2$ is cleared back to empty.

\noindent \textbf{Actions}:
For every new token $x_t$, the agent chooses one of the following actions:
\begin{itemize}
\item \textbf{WAIT}: Instant search is not triggered, and the agent waits for the next token. 
		 
\item \textbf{SEARCH}: Typed query $q'$ is issued to the underlying search system, and new results are retrieved. SEARCH action results in following state transition: $q'_1 =  q'_1 \cup q'_2 \nonumber$ ;  $q'_2=\emptyset \nonumber$

\end{itemize}

\noindent \textbf{Reward}:
During training, at every time-step $t$, the agent receives reward $r_t$ based on $(s_t,a_t)$. 
The reward function is designed to encourage the agent to improve the search result's quality while keeping the number of searches issued to the underlying retrieval engine low. The agent receives a positive reward if a SEARCH (S) action leads to an improvement in Mean Average Precision (MAP) by more than a fixed threshold $R_{th}$. Otherwise, a constant penalty of -1 is imposed. The positive reward is directly proportional to the improvement in map: $\Delta_{MAP}$. We treat the threshold $R_{th}$ as a hyper-parameter and the actual value is later stated in Section \ref{ssec:eval}. Since the WAIT (W) action does not affect the Quality and Total Searches, the reward is set as $0$. The following equation summarizes the reward function:
\begin{gather}
R = 
    \begin{cases}
        $0$, & \text{action = \textbf{W}} \\
        $1 + $\Delta$ MAP$, & \text{action = \textbf{S}  and $\Delta_{MAP} \geq R_{th}$} \\
        $-1$, & \text{action = \textbf{S} and $\Delta_{MAP} < R_{th}$}
    \end{cases} \nonumber
\end{gather}



\begin{figure}[t]
\includegraphics[width=\linewidth]{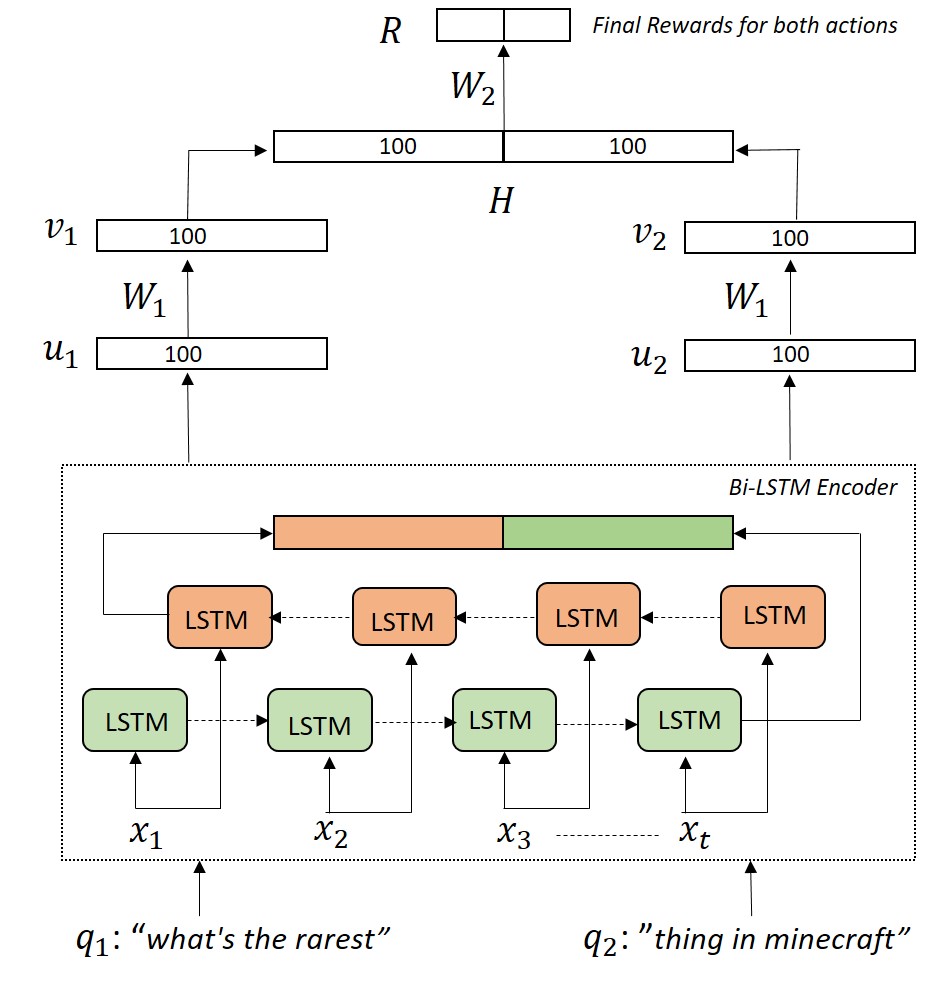}

\caption{The Bi-LSTM Siamese DQN network for calculating rewards for both WAIT and SEARCH actions. The numbers represent the dimension of outputs generated by each layer.}\label{fig:model}
\end{figure}

\noindent \textbf{Bi-LSTM Deep Q Network}:This section describes the base network architecture, as shown in Figure \ref{fig:model} that calculates rewards for a given state. Input to the model is the state, formulated as a pair of sub-queries $(q_1, q_2)$. Input tokens for each sub-query are represented using pre-trained GloVe~\cite{pennington-etal-2014-glove} word-embeddings, that are then passed to a Bi-Directional Long Short-Term Memory (LSTM)~\cite{hochreiter1997long} Siamese Encoder. Since both the sequences have originated from the same query, it is intuitive to apply a Siamese Network that allows the sharing of Bi-LSTM weights. The output vectors for both the sub-queries are concatenated, and the final single feature vector is fed to a fully-connected layer that generates a two-dimensional vector representing the rewards for both the actions. The whole network can be summarized using the below equation:
\setlength{\belowdisplayskip}{0pt} \setlength{\belowdisplayshortskip}{0pt}
\setlength{\abovedisplayskip}{0pt} \setlength{\abovedisplayshortskip}{0pt}
\begin{equation*}
  \begin{gathered}
  u_1 = f_{Bi\mbox{-}LSTM}(q_1)  ;
u_2 = f_{Bi\mbox{-}LSTM}(q_2)  \\
v_1 = relu(W_1{\cdot}u_1 + b_1)  ;
v_2 = relu(W_1{\cdot}u_2 + b_1)  \\
H = [v_1 \oplus v_2]  ;
R = W_2{\cdot}H + b_2 
\end{gathered}
\end{equation*}

\noindent \textbf{Inference}:
For every typed token during an instant search session, a state is prepared as a pair of sub-queries: prefix of the already searched query and suffix that still needs to be searched. The state is passed as an input to the trained model described in Figure \ref{fig:model}. The base model generates rewards for both WAIT and SEARCH actions. The agent picks the action with the best reward, and based on that, the search to the underlying system is either skipped or triggered using the query entered so far. The state for the agent is updated accordingly, and the agent waits for the next token. An episode terminates at the end of the query session.

\section{Metrics and Experimental Setup}
\label{ssec:eval}



\noindent \textbf{Metrics}: We utilize the following metrics to evaluate the performance of the proposed methods.

\noindent \textit{Average Number of Triggered Searches - System Load}:
This metric represents the load on the search system and is measured as the average number of requests made to the search system during an instant search session.

\noindent \textit{Average Effort}:
Studies~\cite{cetindil2012analysis,chandar2019developing} have found the {Effort} to be a very crucial factor that differentiates an instant search user-experience from a traditional search system. {Effort} is defined as the minimum number of tokens that a user would have to type to retrieve the best possible ranking of results. Ranking quality is measured using Mean Average Precision (\textit{MAP}) and the best ranking achieves the maximum MAP. Let $N_q$ be the number of tokens in a given query $q$.
$n_{q}$ is the minimum number of tokens needed to retrieve the best possible ranking for query $q$.
Metric $effort$ is the average effort across all queries in the dataset and is computed as follows:


\setlength{\belowdisplayskip}{0pt} \setlength{\belowdisplayshortskip}{0pt}
\setlength{\abovedisplayskip}{0pt} \setlength{\abovedisplayshortskip}{0pt}
\begin{equation*}
    effort = \frac{\sum n_{q} \leq N_q}{\left | Q \right |} \forall q \in Q
\end{equation*}

\noindent \textit{Quality}: We use MAP to capture the quality of the results. MAP is calculated using the open source PyTREC-Eval \cite{VanGysel2018pytreceval} library.

\noindent \textbf{Evaluation Procedure}:
To measure the TS vs. Effort trade-off, we simulate an action function in a real instant search session for every query and keep track of both the metrics. The action function returns an action($WAIT$ and  $SEARCH$) at every new token based on the decision taken by the method being evaluated. 
For instance, the TS{set}(subsection \ref{SET}) method would return $SEARCH$ for every token in the query. TS is incremented, and results get updated at every search. For every query, we invoke the action function until the retrieval has achieved the best possible MAP or has reached the last token. The total number of tokens used to achieve the best MAP is added to the Effort at the end of the query session. For Quality, we keep track of the MAP achieved at every token position for all the queries. 

\noindent \textbf{Datasets}:We have evaluated the methods on three IR datasets: MS Marco passage ranking\cite{Campos2016MSMA}, Wiki IR\cite{frej2019wikir} 59k version and  InsuranceQA \cite{Feng2015ApplyingDL}. InsuranceQA is adapted to a pure Document Retrieval task using \cite{Tran2018MultihopAN}. InsuranceQA is used in order to test how well methods generalize to different domains. To ensure that the underlying search engine can retrieve relevant documents in top 1000 for enough queries, we have reduced the total number of documents to 400k and 500k for MS Marco and Wiki IR, respectively by random sampling. For InsuranceQA, we use the full set of 27,413 documents. The evaluation sets of size 1000 queries are kept unseen for all three datasets.

\noindent \textbf{Retrievers}: We conduct experiments using both the BM25\cite{Robertson2009ThePR} and semantic-based matching retrieval systems. 
For semantic retrieval, we use a transformer-based pre-trained sentence encoding model known as Universal Sentence Encoder (USE) \cite{cer2018universal} for representing the queries and documents with embeddings and further use cosine similarity to rank results.

\noindent \textbf{Hyper-parameters}:For TS{sm}, we set a threshold of 0.1.
For the proposed TS{dqn} agent, we trained the model with following settings: future reward $\gamma=0.05$, $\epsilon=1$, $\epsilon_{decay} =0.995$, learning rate $\alpha=0.01$ and $\epsilon_{min}=0.7$. 
Furthermore, weights are learned using Adam optimizer\cite{kingma2014adam} with a batch size of 32.
Reward threshold $R_{th}$ mentioned in the reward section for determining the action is set to 0 for MS Marco and Wiki IR and 0.0001 for Insurance QA.

\section{Results}
\begin{table*}[t]
\centering
\tiny
\footnotesize
\resizebox{\textwidth}{!}
{

\begin{tabular}{ccccccc}
\toprule

 & \multicolumn{2}{c}{\textbf{MS Marco - BM25}} & \multicolumn{2}{c}{\textbf{Wiki IR - BM25}} & \multicolumn{2}{c}{\textbf{Insurance QA - USE}} \\
\cmidrule(l){2-3} \cmidrule(l){4-5} \cmidrule(l){6-7} 
\textbf{Methods} & Effort & TS &  Effort & TS &  Effort & TS  \\

\cmidrule(l){1-1} \cmidrule(l){2-3} \cmidrule(l){4-5} \cmidrule(l){6-7} 

{SLT (Regular Search)} 
& 10.76 & 1 &  5.83 & 1 & 8.25 & 1 \\
{SET (Instant Search)} 
& 8.24 & 8.24 & 4.74 & 4.74 & 7.70
 & 7.70  \\
 \cmidrule(l){1-7} 
 \multicolumn{7}{c}{Percentage change in metrics with respect to SET(Pure Instant Search)} \\
 \cmidrule(l){1-7} 
 \cmidrule(l){1-1} \cmidrule(l){2-3} \cmidrule(l){4-5} \cmidrule(l){6-7} 
 & $\Delta$Effort(\%) & $\Delta$TS(\%) &  $\Delta$Effort(\%) & $\Delta$TS(\%) &  $\Delta$Effort(\%) & $\Delta$TS(\%)  \\
\cmidrule(l){1-1} \cmidrule(l){2-3} \cmidrule(l){4-5} \cmidrule(l){6-7} 

{SS (Baseline)} 
&  0 & -49.75  & 
 0 & -22.62 &
 0.59 & -39.25
 \\
 
{SM (Baseline)} 
& 4.00 & -45.43 & 3.24 & -26.88   & 1.50 & -40.42  \\

{DQN (Proposed)} 
&  4.00 & \textbf{-74.15}\textsuperscript{*}
 & 3.94 & \textbf{-44.88}\textsuperscript{*}
 & 1.37 & \textbf{-55.47}\textsuperscript{*}  \\


\bottomrule
\end{tabular}
}
\caption{Metrics achieved by different methods. Effort and TS metrics are averaged over all the queries. The top two rows are the absolute values achieved by two base search systems. The bottom three rows list down the \% change in the metrics introduced by methods with respect to a true instant search system. \textsuperscript{*}Statistical significance is tested using a two-tailed paired t-test. We mark significant improvements when p $<$ 0.01}

\label{tbl:results}
\end{table*}

\begin{figure*}[t]
\centering
    \includegraphics[width=\linewidth]{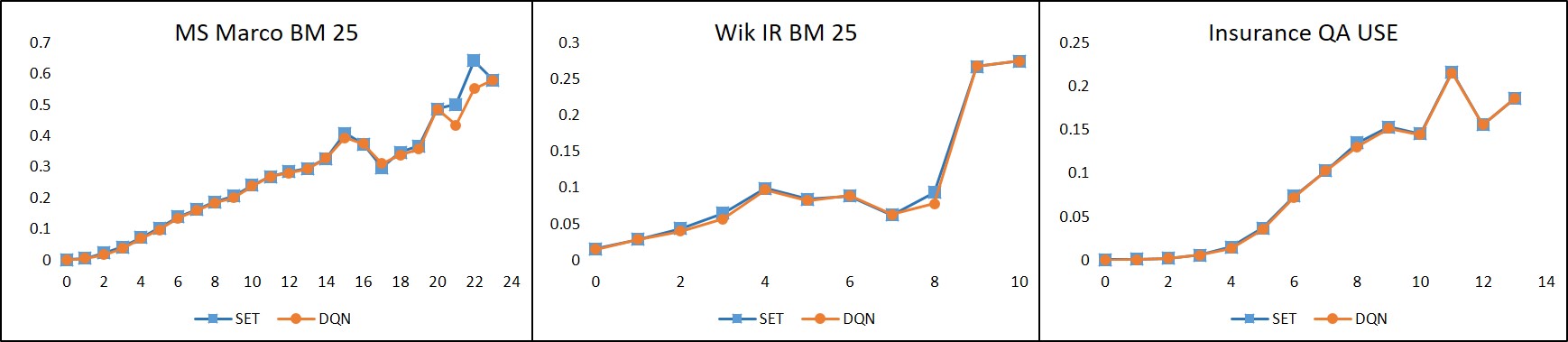}
    \caption{Average MAP achieved by DQN vs. SET at every token position. X-axis is token index and Y-axis is MAP averaged over all the queries.}
    \label{fig:line_chart}
\end{figure*}
\begin{figure*}[t]
\centering
    \includegraphics[width=\linewidth]{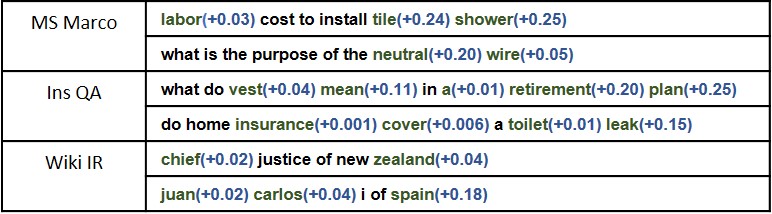}
    \caption{Predictions of DQN Network on unseen queries.}
    \label{fig:sayt-queries}
    
\end{figure*}

\noindent \textbf{TS vs. Effort}: Table \ref{tbl:results} shows the drop in Average Number of Triggered Searches achieved by different methods and compares it with extra Average Effort introduced in the system. The top two rows highlight the absolute values achieved by two basic search systems: SLT, which mimics a traditional search engine, and SET, representing a true instant search system.  These systems set the upper and lower bounds respectively on Effort and TS. The bottom three rows list down the percentage change in the metrics introduced by the proposed methods with respect to a true instant search system(SET). 


Skip Stop-words method manages to achieve optimum Effort.
This can be attributed to the fact that generally, stop-words are not deemed salient in common language usage, allowing SS not to miss a search for any salient words.
On the other hand, since SS influences only a limited and fixed set of tokens, the achieved TS is not up to the optimal. Also, the overall performance of SM is quite comparable to SS. Results also show that that the pre-trained model is unable to transfer its knowledge to this new task. 

While all the methods are able to retain the Effort within 5\% of the SET, the proposed DQN method manages to reduce the overall TS on average by more than 20\% across all the datasets compared to other baselines. Compared to a pure instant search system, DQN reduces the overall load by more than 50\%. The performance of DQN agent is directly proportional to the training size of the dataset and hence is highest for MS Marco.

\noindent \textbf{Impact on Quality}: We have captured the loss of quality in results at every token position by plotting the average MAP over all the queries at every token position for the proposed DQN method and further comparing it with the ideal SET instant search. Figure \ref{fig:line_chart} plots the average MAP(y-axis) at every time-step $t$(x-axis) for both SET and DQN. The plot shows that the MAP achieved by DQN is very close to that of SET at all the token positions, and hence the loss in quality introduced is minimal.
\noindent \textbf{Subjective Analysis}: Lastly, we subjectively analyzed the predictions made on the unseen queries by the learned model. Figure \ref{fig:sayt-queries} lists queries with tokens at which the search triggered is marked green. We also report the incremental difference in MAP introduced by the triggered search(highlighted as blue) to capture the search action quality. 

For MS Marco, besides stop-words, the agent waits for the words \textit{``cost'' }, \textit{``install''} and \textit{``purpose''}.  MS Marco is a large QA dataset with verbose passages. It is difficult for a basic BM25 algorithm to retrieve a good ranking without additional context early in the query; thus the model decides to wait. For the first InsuranceQA query, the agent decides to execute the search for the token \textit{``a''} as in insurance jargon, ``vest a retirement plan'' is a common phrase, and a semantic model such as Universal Sentence Encoder does not ignore this as a stop-word. The same is not true for the second sentence where the search is skipped for the same token. For WikiIR's first query, the phrase \textit{``chief justice''} is often present as a whole in documents, and any improvement in ranking contributed by the phrase itself is already captured by the first word \textit{``chief''}. In the second example, the name \textit{``juan carlos''} is unique enough to retrieve relevant documents; therefore, the token \textit{``i''} is skipped.
\section{Conclusion}
\vspace{-1mm}
This paper has introduced a Reinforcement Agent that relieves the load on the back-end search system in an instant search paradigm. Proposed agent achieves the goal by learning word importance based on the search system behavior and utilizes this knowledge towards judiciously issuing searches to the underlying retrieval system. We further evaluated the trade-off between system load and performance. Experiments demonstrate the ability of the proposed agent to achieve near-optimal trade-off. 

\bibliographystyle{ACM-Reference-Format}
\bibliography{sayt-citations.bib}


\begin{thebibliography}{35}


\ifx \showCODEN    \undefined \def \showCODEN     #1{\unskip}     \fi
\ifx \showDOI      \undefined \def \showDOI       #1{#1}\fi
\ifx \showISBNx    \undefined \def \showISBNx     #1{\unskip}     \fi
\ifx \showISBNxiii \undefined \def \showISBNxiii  #1{\unskip}     \fi
\ifx \showISSN     \undefined \def \showISSN      #1{\unskip}     \fi
\ifx \showLCCN     \undefined \def \showLCCN      #1{\unskip}     \fi
\ifx \shownote     \undefined \def \shownote      #1{#1}          \fi
\ifx \showarticletitle \undefined \def \showarticletitle #1{#1}   \fi
\ifx \showURL      \undefined \def \showURL       {\relax}        \fi
\providecommand\bibfield[2]{#2}
\providecommand\bibinfo[2]{#2}
\providecommand\natexlab[1]{#1}
\providecommand\showeprint[2][]{arXiv:#2}

\bibitem[\protect\citeauthoryear{Bast and Weber}{Bast and Weber}{2006}]%
        {Bast2006TypeLF}
\bibfield{author}{\bibinfo{person}{Holger Bast} {and} \bibinfo{person}{Ingmar
  Weber}.} \bibinfo{year}{2006}\natexlab{}.
\newblock \showarticletitle{Type less, find more: fast autocompletion search
  with a succinct index}. In \bibinfo{booktitle}{\emph{Proceedings of the 29th
  annual international ACM SIGIR conference on Research and development in
  information retrieval}}. \bibinfo{pages}{364--371}.
\newblock


\bibitem[\protect\citeauthoryear{Brin and Page}{Brin and Page}{1998}]%
        {brin1998anatomy}
\bibfield{author}{\bibinfo{person}{Sergey Brin} {and} \bibinfo{person}{Lawrence
  Page}.} \bibinfo{year}{1998}\natexlab{}.
\newblock \showarticletitle{The anatomy of a large-scale hypertextual web
  search engine}.
\newblock  (\bibinfo{year}{1998}).
\newblock


\bibitem[\protect\citeauthoryear{Cer, Yang, Kong, Hua, Limtiaco, John,
  Constant, Guajardo-Cespedes, Yuan, Tar, et~al\mbox{.}}{Cer
  et~al\mbox{.}}{2018}]%
        {cer2018universal}
\bibfield{author}{\bibinfo{person}{Daniel Cer}, \bibinfo{person}{Yinfei Yang},
  \bibinfo{person}{Sheng-yi Kong}, \bibinfo{person}{Nan Hua},
  \bibinfo{person}{Nicole Limtiaco}, \bibinfo{person}{Rhomni~St John},
  \bibinfo{person}{Noah Constant}, \bibinfo{person}{Mario Guajardo-Cespedes},
  \bibinfo{person}{Steve Yuan}, \bibinfo{person}{Chris Tar}, {et~al\mbox{.}}}
  \bibinfo{year}{2018}\natexlab{}.
\newblock \showarticletitle{Universal sentence encoder}.
\newblock \bibinfo{journal}{\emph{arXiv preprint arXiv:1803.11175}}
  (\bibinfo{year}{2018}).
\newblock


\bibitem[\protect\citeauthoryear{Cetindil, Esmaelnezhad, Li, and
  Newman}{Cetindil et~al\mbox{.}}{2012}]%
        {cetindil2012analysis}
\bibfield{author}{\bibinfo{person}{Inci Cetindil}, \bibinfo{person}{Jamshid
  Esmaelnezhad}, \bibinfo{person}{Chen Li}, {and} \bibinfo{person}{David
  Newman}.} \bibinfo{year}{2012}\natexlab{}.
\newblock \showarticletitle{Analysis of Instant Search Query Logs.}. In
  \bibinfo{booktitle}{\emph{WebDB}}. Citeseer, \bibinfo{pages}{7--12}.
\newblock


\bibitem[\protect\citeauthoryear{Chandar, Garcia-Gathright, Hosey, St.~Thomas,
  and Thom}{Chandar et~al\mbox{.}}{2019}]%
        {chandar2019developing}
\bibfield{author}{\bibinfo{person}{Praveen Chandar}, \bibinfo{person}{Jean
  Garcia-Gathright}, \bibinfo{person}{Christine Hosey}, \bibinfo{person}{Brian
  St.~Thomas}, {and} \bibinfo{person}{Jennifer Thom}.}
  \bibinfo{year}{2019}\natexlab{}.
\newblock \showarticletitle{Developing Evaluation Metrics for Instant Search
  Using Mixed Methods Methods}. In \bibinfo{booktitle}{\emph{Proceedings of the
  42nd International ACM SIGIR Conference on Research and Development in
  Information Retrieval}}. \bibinfo{pages}{925--928}.
\newblock


\bibitem[\protect\citeauthoryear{Dean}{Dean}{2009}]%
        {dean2009challenges}
\bibfield{author}{\bibinfo{person}{Jeffrey Dean}.}
  \bibinfo{year}{2009}\natexlab{}.
\newblock \showarticletitle{Challenges in building large-scale information
  retrieval systems}. In \bibinfo{booktitle}{\emph{Keynote of the 2nd ACM
  International Conference on Web Search and Data Mining (WSDM)}},
  Vol.~\bibinfo{volume}{10}.
\newblock


\bibitem[\protect\citeauthoryear{Fafalios, Kitsos, and Tzitzikas}{Fafalios
  et~al\mbox{.}}{2012}]%
        {Fafalios2012ScalableFA}
\bibfield{author}{\bibinfo{person}{Pavlos Fafalios}, \bibinfo{person}{Ioannis
  Kitsos}, {and} \bibinfo{person}{Yannis Tzitzikas}.}
  \bibinfo{year}{2012}\natexlab{}.
\newblock \showarticletitle{Scalable, flexible and generic instant overview
  search}. In \bibinfo{booktitle}{\emph{Proceedings of the 21st International
  Conference on World Wide Web}}. \bibinfo{pages}{333--336}.
\newblock


\bibitem[\protect\citeauthoryear{Fafalios and Tzitzikas}{Fafalios and
  Tzitzikas}{2011}]%
        {Fafalios2011ExploitingAM}
\bibfield{author}{\bibinfo{person}{Pavlos Fafalios} {and}
  \bibinfo{person}{Yannis Tzitzikas}.} \bibinfo{year}{2011}\natexlab{}.
\newblock \showarticletitle{Exploiting available memory and disk for scalable
  instant overview search}. In \bibinfo{booktitle}{\emph{International
  Conference on Web Information Systems Engineering}}. Springer,
  \bibinfo{pages}{101--115}.
\newblock


\bibitem[\protect\citeauthoryear{Fagni, Perego, Silvestri, and Orlando}{Fagni
  et~al\mbox{.}}{2006}]%
        {fagni2006boosting}
\bibfield{author}{\bibinfo{person}{Tiziano Fagni}, \bibinfo{person}{Raffaele
  Perego}, \bibinfo{person}{Fabrizio Silvestri}, {and}
  \bibinfo{person}{Salvatore Orlando}.} \bibinfo{year}{2006}\natexlab{}.
\newblock \showarticletitle{Boosting the performance of web search engines:
  Caching and prefetching query results by exploiting historical usage data}.
\newblock \bibinfo{journal}{\emph{ACM Transactions on Information Systems
  (TOIS)}} \bibinfo{volume}{24}, \bibinfo{number}{1} (\bibinfo{year}{2006}),
  \bibinfo{pages}{51--78}.
\newblock


\bibitem[\protect\citeauthoryear{Feng, Xiang, Glass, Wang, and Zhou}{Feng
  et~al\mbox{.}}{2015}]%
        {Feng2015ApplyingDL}
\bibfield{author}{\bibinfo{person}{Minwei Feng}, \bibinfo{person}{Bing Xiang},
  \bibinfo{person}{Michael~R Glass}, \bibinfo{person}{Lidan Wang}, {and}
  \bibinfo{person}{Bowen Zhou}.} \bibinfo{year}{2015}\natexlab{}.
\newblock \showarticletitle{Applying deep learning to answer selection: A study
  and an open task}. In \bibinfo{booktitle}{\emph{2015 IEEE Workshop on
  Automatic Speech Recognition and Understanding (ASRU)}}. IEEE,
  \bibinfo{pages}{813--820}.
\newblock


\bibitem[\protect\citeauthoryear{Frej, Schwab, and Chevallet}{Frej
  et~al\mbox{.}}{2019}]%
        {frej2019wikir}
\bibfield{author}{\bibinfo{person}{Jibril Frej}, \bibinfo{person}{Didier
  Schwab}, {and} \bibinfo{person}{Jean-Pierre Chevallet}.}
  \bibinfo{year}{2019}\natexlab{}.
\newblock \showarticletitle{WIKIR: A Python toolkit for building a large-scale
  Wikipedia-based English Information Retrieval Dataset}.
\newblock \bibinfo{journal}{\emph{arXiv preprint arXiv:1912.01901}}
  (\bibinfo{year}{2019}).
\newblock


\bibitem[\protect\citeauthoryear{Gan and Suel}{Gan and Suel}{2009}]%
        {gan2009improved}
\bibfield{author}{\bibinfo{person}{Qingqing Gan} {and} \bibinfo{person}{Torsten
  Suel}.} \bibinfo{year}{2009}\natexlab{}.
\newblock \showarticletitle{Improved techniques for result caching in web
  search engines}. In \bibinfo{booktitle}{\emph{Proceedings of the 18th
  international conference on World wide web}}. \bibinfo{pages}{431--440}.
\newblock


\bibitem[\protect\citeauthoryear{Gomaa, Fahmy, et~al\mbox{.}}{Gomaa
  et~al\mbox{.}}{2013}]%
        {gomaa2013survey}
\bibfield{author}{\bibinfo{person}{Wael~H Gomaa}, \bibinfo{person}{Aly~A
  Fahmy}, {et~al\mbox{.}}} \bibinfo{year}{2013}\natexlab{}.
\newblock \showarticletitle{A survey of text similarity approaches}.
\newblock \bibinfo{journal}{\emph{International Journal of Computer
  Applications}} \bibinfo{volume}{68}, \bibinfo{number}{13}
  (\bibinfo{year}{2013}), \bibinfo{pages}{13--18}.
\newblock


\bibitem[\protect\citeauthoryear{Grissom~II, He, Boyd-Graber, Morgan, and
  Daum{\'e}~III}{Grissom~II et~al\mbox{.}}{2014}]%
        {grissom}
\bibfield{author}{\bibinfo{person}{Alvin Grissom~II}, \bibinfo{person}{He He},
  \bibinfo{person}{Jordan Boyd-Graber}, \bibinfo{person}{John Morgan}, {and}
  \bibinfo{person}{Hal Daum{\'e}~III}.} \bibinfo{year}{2014}\natexlab{}.
\newblock \showarticletitle{Don’t until the final verb wait: Reinforcement
  learning for simultaneous machine translation}. In
  \bibinfo{booktitle}{\emph{Proceedings of the 2014 Conference on empirical
  methods in natural language processing (EMNLP)}}.
  \bibinfo{pages}{1342--1352}.
\newblock


\bibitem[\protect\citeauthoryear{Gu, Neubig, Cho, and Li}{Gu
  et~al\mbox{.}}{2016}]%
        {Neubig2017LearningTT}
\bibfield{author}{\bibinfo{person}{Jiatao Gu}, \bibinfo{person}{Graham Neubig},
  \bibinfo{person}{Kyunghyun Cho}, {and} \bibinfo{person}{Victor~OK Li}.}
  \bibinfo{year}{2016}\natexlab{}.
\newblock \showarticletitle{Learning to translate in real-time with neural
  machine translation}.
\newblock \bibinfo{journal}{\emph{arXiv preprint arXiv:1610.00388}}
  (\bibinfo{year}{2016}).
\newblock


\bibitem[\protect\citeauthoryear{Hochreiter and Schmidhuber}{Hochreiter and
  Schmidhuber}{1997}]%
        {hochreiter1997long}
\bibfield{author}{\bibinfo{person}{Sepp Hochreiter} {and}
  \bibinfo{person}{J{\"u}rgen Schmidhuber}.} \bibinfo{year}{1997}\natexlab{}.
\newblock \showarticletitle{Long short-term memory}.
\newblock \bibinfo{journal}{\emph{Neural computation}} \bibinfo{volume}{9},
  \bibinfo{number}{8} (\bibinfo{year}{1997}), \bibinfo{pages}{1735--1780}.
\newblock


\bibitem[\protect\citeauthoryear{Ji, Li, Li, and Feng}{Ji
  et~al\mbox{.}}{2009}]%
        {ji2009efficient}
\bibfield{author}{\bibinfo{person}{Shengyue Ji}, \bibinfo{person}{Guoliang Li},
  \bibinfo{person}{Chen Li}, {and} \bibinfo{person}{Jianhua Feng}.}
  \bibinfo{year}{2009}\natexlab{}.
\newblock \showarticletitle{Efficient interactive fuzzy keyword search}. In
  \bibinfo{booktitle}{\emph{Proceedings of the 18th international conference on
  World wide web}}. \bibinfo{pages}{371--380}.
\newblock


\bibitem[\protect\citeauthoryear{Kingma and Ba}{Kingma and Ba}{2014}]%
        {kingma2014adam}
\bibfield{author}{\bibinfo{person}{Diederik~P Kingma} {and}
  \bibinfo{person}{Jimmy Ba}.} \bibinfo{year}{2014}\natexlab{}.
\newblock \showarticletitle{Adam: A method for stochastic optimization}.
\newblock \bibinfo{journal}{\emph{arXiv preprint arXiv:1412.6980}}
  (\bibinfo{year}{2014}).
\newblock


\bibitem[\protect\citeauthoryear{Li, Ji, Li, and Feng}{Li
  et~al\mbox{.}}{2011}]%
        {Li2011EfficientFF}
\bibfield{author}{\bibinfo{person}{Guoliang Li}, \bibinfo{person}{Shengyue Ji},
  \bibinfo{person}{Chen Li}, {and} \bibinfo{person}{Jianhua Feng}.}
  \bibinfo{year}{2011}\natexlab{}.
\newblock \showarticletitle{Efficient fuzzy full-text type-ahead search}.
\newblock \bibinfo{journal}{\emph{The VLDB Journal}} \bibinfo{volume}{20},
  \bibinfo{number}{4} (\bibinfo{year}{2011}), \bibinfo{pages}{617--640}.
\newblock


\bibitem[\protect\citeauthoryear{Li, Wang, Li, and Feng}{Li
  et~al\mbox{.}}{2012}]%
        {Li2012SupportingET}
\bibfield{author}{\bibinfo{person}{Guoliang Li}, \bibinfo{person}{Jiannan
  Wang}, \bibinfo{person}{Chen Li}, {and} \bibinfo{person}{Jianhua Feng}.}
  \bibinfo{year}{2012}\natexlab{}.
\newblock \showarticletitle{Supporting efficient top-k queries in type-ahead
  search}. In \bibinfo{booktitle}{\emph{Proceedings of the 35th international
  ACM SIGIR conference on Research and development in information retrieval}}.
  \bibinfo{pages}{355--364}.
\newblock


\bibitem[\protect\citeauthoryear{Long and Suel}{Long and Suel}{2006}]%
        {long2006three}
\bibfield{author}{\bibinfo{person}{Xiaohui Long} {and} \bibinfo{person}{Torsten
  Suel}.} \bibinfo{year}{2006}\natexlab{}.
\newblock \showarticletitle{Three-level caching for efficient query processing
  in large web search engines}.
\newblock \bibinfo{journal}{\emph{World Wide Web}} \bibinfo{volume}{9},
  \bibinfo{number}{4} (\bibinfo{year}{2006}), \bibinfo{pages}{369--395}.
\newblock


\bibitem[\protect\citeauthoryear{Markatos}{Markatos}{2001}]%
        {MARKATOS2001137}
\bibfield{author}{\bibinfo{person}{Evangelos~P. Markatos}.}
  \bibinfo{year}{2001}\natexlab{}.
\newblock \showarticletitle{On caching search engine query results}.
\newblock \bibinfo{journal}{\emph{Computer Communications}}
  \bibinfo{volume}{24}, \bibinfo{number}{2} (\bibinfo{year}{2001}),
  \bibinfo{pages}{137--143}.
\newblock


\bibitem[\protect\citeauthoryear{Mnih, Kavukcuoglu, Silver, Graves, Antonoglou,
  Wierstra, and Riedmiller}{Mnih et~al\mbox{.}}{2013}]%
        {Mnih2013PlayingAW}
\bibfield{author}{\bibinfo{person}{Volodymyr Mnih}, \bibinfo{person}{Koray
  Kavukcuoglu}, \bibinfo{person}{David Silver}, \bibinfo{person}{Alex Graves},
  \bibinfo{person}{Ioannis Antonoglou}, \bibinfo{person}{Daan Wierstra}, {and}
  \bibinfo{person}{Martin Riedmiller}.} \bibinfo{year}{2013}\natexlab{}.
\newblock \showarticletitle{Playing atari with deep reinforcement learning}.
\newblock \bibinfo{journal}{\emph{arXiv preprint arXiv:1312.5602}}
  (\bibinfo{year}{2013}).
\newblock


\bibitem[\protect\citeauthoryear{Nguyen, Rosenberg, Song, Gao, Tiwary,
  Majumder, and Deng}{Nguyen et~al\mbox{.}}{2016}]%
        {Campos2016MSMA}
\bibfield{author}{\bibinfo{person}{Tri Nguyen}, \bibinfo{person}{Mir
  Rosenberg}, \bibinfo{person}{Xia Song}, \bibinfo{person}{Jianfeng Gao},
  \bibinfo{person}{Saurabh Tiwary}, \bibinfo{person}{Rangan Majumder}, {and}
  \bibinfo{person}{Li Deng}.} \bibinfo{year}{2016}\natexlab{}.
\newblock \showarticletitle{Ms marco: A human-generated machine reading
  comprehension dataset}.
\newblock  (\bibinfo{year}{2016}).
\newblock


\bibitem[\protect\citeauthoryear{Pennington, Socher, and Manning}{Pennington
  et~al\mbox{.}}{2014}]%
        {pennington-etal-2014-glove}
\bibfield{author}{\bibinfo{person}{Jeffrey Pennington},
  \bibinfo{person}{Richard Socher}, {and} \bibinfo{person}{Christopher~D
  Manning}.} \bibinfo{year}{2014}\natexlab{}.
\newblock \showarticletitle{Glove: Global vectors for word representation}. In
  \bibinfo{booktitle}{\emph{Proceedings of the 2014 conference on empirical
  methods in natural language processing (EMNLP)}}.
  \bibinfo{pages}{1532--1543}.
\newblock


\bibitem[\protect\citeauthoryear{Robertson and Zaragoza}{Robertson and
  Zaragoza}{2009}]%
        {Robertson2009ThePR}
\bibfield{author}{\bibinfo{person}{Stephen Robertson} {and}
  \bibinfo{person}{Hugo Zaragoza}.} \bibinfo{year}{2009}\natexlab{}.
\newblock \bibinfo{booktitle}{\emph{The probabilistic relevance framework: BM25
  and beyond}}.
\newblock \bibinfo{publisher}{Now Publishers Inc}.
\newblock


\bibitem[\protect\citeauthoryear{Saraiva, Silva~de Moura, Ziviani, Meira,
  Fonseca, and Ribeiro-Neto}{Saraiva et~al\mbox{.}}{2001}]%
        {saraiva2001rank}
\bibfield{author}{\bibinfo{person}{Patricia~Correia Saraiva},
  \bibinfo{person}{Edleno Silva~de Moura}, \bibinfo{person}{Nivio Ziviani},
  \bibinfo{person}{Wagner Meira}, \bibinfo{person}{Rodrigo Fonseca}, {and}
  \bibinfo{person}{Berthier Ribeiro-Neto}.} \bibinfo{year}{2001}\natexlab{}.
\newblock \showarticletitle{Rank-preserving two-level caching for scalable
  search engines}. In \bibinfo{booktitle}{\emph{Proceedings of the 24th annual
  international ACM SIGIR conference on Research and development in information
  retrieval}}. \bibinfo{pages}{51--58}.
\newblock


\bibitem[\protect\citeauthoryear{Satija and Pineau}{Satija and Pineau}{2016}]%
        {satija2016simultaneous}
\bibfield{author}{\bibinfo{person}{Harsh Satija} {and} \bibinfo{person}{Joelle
  Pineau}.} \bibinfo{year}{2016}\natexlab{}.
\newblock \showarticletitle{Simultaneous machine translation using deep
  reinforcement learning}. In \bibinfo{booktitle}{\emph{ICML 2016 Workshop on
  Abstraction in Reinforcement Learning}}.
\newblock


\bibitem[\protect\citeauthoryear{Tran and Niedere{\'e}e}{Tran and
  Niedere{\'e}e}{2018}]%
        {Tran2018MultihopAN}
\bibfield{author}{\bibinfo{person}{Nam~Khanh Tran} {and}
  \bibinfo{person}{Claudia Niedere{\'e}e}.} \bibinfo{year}{2018}\natexlab{}.
\newblock \showarticletitle{Multihop attention networks for question answer
  matching}. In \bibinfo{booktitle}{\emph{The 41st International ACM SIGIR
  Conference on Research \& Development in Information Retrieval}}.
  \bibinfo{pages}{325--334}.
\newblock


\bibitem[\protect\citeauthoryear{Van~Gysel and de~Rijke}{Van~Gysel and
  de~Rijke}{2018}]%
        {VanGysel2018pytreceval}
\bibfield{author}{\bibinfo{person}{Christophe Van~Gysel} {and}
  \bibinfo{person}{Maarten de Rijke}.} \bibinfo{year}{2018}\natexlab{}.
\newblock \showarticletitle{Pytrec\_eval: An extremely fast python interface to
  trec\_eval}. In \bibinfo{booktitle}{\emph{The 41st International ACM SIGIR
  Conference on Research \& Development in Information Retrieval}}.
  \bibinfo{pages}{873--876}.
\newblock


\bibitem[\protect\citeauthoryear{Venkataraman, Lad, Guo, and
  Sinha}{Venkataraman et~al\mbox{.}}{2016a}]%
        {venkataraman2016fast}
\bibfield{author}{\bibinfo{person}{Ganesh Venkataraman},
  \bibinfo{person}{Abhimanyu Lad}, \bibinfo{person}{Lin Guo}, {and}
  \bibinfo{person}{Shakti Sinha}.} \bibinfo{year}{2016}\natexlab{a}.
\newblock \showarticletitle{Fast, lenient and accurate: Building personalized
  instant search experience at linkedin}. In \bibinfo{booktitle}{\emph{2016
  IEEE International Conference on Big Data (Big Data)}}. IEEE,
  \bibinfo{pages}{1502--1511}.
\newblock


\bibitem[\protect\citeauthoryear{Venkataraman, Lad, Ha-Thuc, and
  Arya}{Venkataraman et~al\mbox{.}}{2016b}]%
        {venkataraman2016instant}
\bibfield{author}{\bibinfo{person}{Ganesh Venkataraman},
  \bibinfo{person}{Abhimanyu Lad}, \bibinfo{person}{Viet Ha-Thuc}, {and}
  \bibinfo{person}{Dhruv Arya}.} \bibinfo{year}{2016}\natexlab{b}.
\newblock \showarticletitle{Instant search: A hands-on tutorial}. In
  \bibinfo{booktitle}{\emph{Proceedings of the 39th International ACM SIGIR
  conference on Research and Development in Information Retrieval}}.
  \bibinfo{pages}{1211--1214}.
\newblock


\bibitem[\protect\citeauthoryear{Wang, Cetindil, Ji, Li, Xie, Li, and
  Feng}{Wang et~al\mbox{.}}{2010}]%
        {Wang2010InteractiveAF}
\bibfield{author}{\bibinfo{person}{Jiannan Wang}, \bibinfo{person}{Inci
  Cetindil}, \bibinfo{person}{Shengyue Ji}, \bibinfo{person}{Chen Li},
  \bibinfo{person}{Xiaohui Xie}, \bibinfo{person}{Guoliang Li}, {and}
  \bibinfo{person}{Jianhua Feng}.} \bibinfo{year}{2010}\natexlab{}.
\newblock \showarticletitle{Interactive and fuzzy search: a dynamic way to
  explore MEDLINE}.
\newblock \bibinfo{journal}{\emph{Bioinformatics}} \bibinfo{volume}{26},
  \bibinfo{number}{18} (\bibinfo{year}{2010}), \bibinfo{pages}{2321--2327}.
\newblock


\bibitem[\protect\citeauthoryear{Watkins and Dayan}{Watkins and Dayan}{1992}]%
        {watkins1992q}
\bibfield{author}{\bibinfo{person}{Christopher~JCH Watkins} {and}
  \bibinfo{person}{Peter Dayan}.} \bibinfo{year}{1992}\natexlab{}.
\newblock \showarticletitle{Q-learning}.
\newblock \bibinfo{journal}{\emph{Machine learning}} \bibinfo{volume}{8},
  \bibinfo{number}{3-4} (\bibinfo{year}{1992}), \bibinfo{pages}{279--292}.
\newblock


\bibitem[\protect\citeauthoryear{Yang, Xie, Lin, Li, Tan, Xiong, Li, and
  Lin}{Yang et~al\mbox{.}}{2019}]%
        {yang2019end}
\bibfield{author}{\bibinfo{person}{Wei Yang}, \bibinfo{person}{Yuqing Xie},
  \bibinfo{person}{Aileen Lin}, \bibinfo{person}{Xingyu Li},
  \bibinfo{person}{Luchen Tan}, \bibinfo{person}{Kun Xiong},
  \bibinfo{person}{Ming Li}, {and} \bibinfo{person}{Jimmy Lin}.}
  \bibinfo{year}{2019}\natexlab{}.
\newblock \showarticletitle{End-to-end open-domain question answering with
  bertserini}.
\newblock \bibinfo{journal}{\emph{arXiv preprint arXiv:1902.01718}}
  (\bibinfo{year}{2019}).
\newblock


\end{thebibliography}
\end{document}